%% file: main.tex
\pdfoutput=1

\documentclass[11pt]{article}

\usepackage{acl}

\usepackage{times}
\usepackage{latexsym}

\usepackage[T1]{fontenc}

\usepackage[utf8]{inputenc}

\usepackage{microtype}

\usepackage{inconsolata}

\usepackage{hyperref}       
\usepackage{url}            
\usepackage{booktabs}       
\usepackage{amsfonts}       
\usepackage{nicefrac}       
\usepackage{microtype}      
\usepackage{xcolor}         
\usepackage{algorithm}
\usepackage{algorithmic}
\usepackage[fleqn]{amsmath}
\usepackage{caption}
\usepackage{subcaption}
\usepackage{gensymb}
\usepackage{amssymb,latexsym}
\usepackage{wrapfig}
\usepackage{todonotes}
\usepackage{tabularx} 
\usepackage{multirow}

\usepackage{natbib} 
\usepackage{makecell} 

\usepackage{libertine}
\usepackage[T1]{fontenc}

\usepackage{fullpage}   

\usepackage{titlesec}
\usepackage{tabto}
\usepackage{amsmath}
\usepackage{lscape}
\usepackage{xcolor}
\usepackage{hhline}
\usepackage{multirow}
\usepackage{booktabs}
\usepackage{arydshln}
\usepackage{siunitx}
\usepackage{footmisc}
\usepackage{pifont}
\usepackage{comment}
\usepackage{blindtext}

\usepackage{url}
\usepackage[normalem]{ulem}
\usepackage[titletoc,title]{appendix}
\usepackage{graphicx}
\usepackage{wrapfig,lipsum}
\usepackage{makecell}
\usepackage{enumitem}
\usepackage{caption}
\usepackage{subcaption}
\usepackage{bm}
\usepackage{amsmath,amsfonts,amssymb,bbm,epsfig,amsthm}
\usepackage{latexsym}

\input{math_commands}

\newcommand{\leqnomode}{\tagsleft@true\let\veqno\@@leqno}%
\newcommand{\reqnomode}{\tagsleft@false\let\veqno\@@eqno}%
\newcommand*{\compress}{\@minipagetrue}

\definecolor{yy}{RGB}{220,220,0}
\definecolor{gg}{RGB}{45,190,45}

\makeatletter
\def\BState{\State\hskip-\ALG@thistlm}
\makeatother

\title{Task-Based MoE for Multitask Multilingual Machine Translation}

\author{
Hai Pham \\
  Carnegie Mellon University \\
  \small\texttt{htpham@cs.cmu.edu} \\
  \And
Young Jin Kim \\
  Microsoft \\
  \small\texttt{ youki@microsoft.com} \\ 
  \And
  Subhabrata Mukherjee\thanks{\,\, Work done while at Microsoft (contact email: \small\texttt{subhabrata.mukherjee.ju@gmail.com}).} \\
  Hippocratic AI \\
  \AND 
David P. Woodruff \\ 
  Carnegie Mellon University \\
  \small\texttt{dwoodruf@cs.cmu.edu} \\ 
  \And 
Barnab\'{a}s P\'{o}czos  \\
  Carnegie Mellon University \\
  \small\texttt{bapoczos@cs.cmu.edu} \\ 
  \And
Hany Hassan Awadalla \\
    Microsoft \\  
  \small\texttt{hanyh@microsoft.com} 
}

\begin{document}
\maketitle

\begin{abstract}
Mixture-of-experts (MoE) architecture has been proven a powerful method for diverse tasks in training deep models in many applications. 
However, current MoE implementations are task agnostic, treating all tokens from different tasks in the same manner. 
In this work, we instead design a novel method that incorporates task information into MoE models at different granular levels with shared dynamic task-based adapters. Our experiments and analysis show the advantages of our approaches over the dense and canonical MoE models on multitask multilingual machine translations. With task-specific adapters, our models can additionally generalize to new tasks efficiently. 
\end{abstract}

\section{Introduction}


Mixture-of-Experts (MoE), while not being a novel machine learning algorithm~\cite{Yksel2012TwentyYO}, has revived to combine with deep learning, particularly transformer~\cite{vaswani2017attention} and has recently pushed forward various tasks such as natural language processing, computer vision, speech recognition, multimodal and multitask learning due to its advantage in scalability in distributed environments \cite{Fedus2022ARO}. 
The main advantages of MoE stem from its ensemble design while maintaining the sparsity in computation~\cite{fedus2021switch}. And with proper design such as using sharded experts~\cite{lepikhin2020gshard,fedus2021switch}, the possibility for enterprise-level scalability is almost boundless. 
As a result, this method has been more and more widely adopted in many applications that require distributed and intensive workloads.

However, most of the current methods are task-agnostic, only optimizing for performance based on lower levels in the architecture such as at system or communication levels~\cite{Rajbhandari2022DeepSpeedMoEAM}. 
In the case of multitask learning where a single model is required to learn from heterogeneous tasks, however, the task-specific data could be inherently diverse and vary largely from one to another~\cite{Wu2020UnderstandingAI}. 
As a result, treating data from such different sources the same makes the learning ineffective, as also evidenced recently by the interference between different task data~\cite{Pfeiffer2022LiftingTC}.

\begin{figure}[t]
    \centering
\label{fig:len_distribution}
  \includegraphics[width=0.45\textwidth]{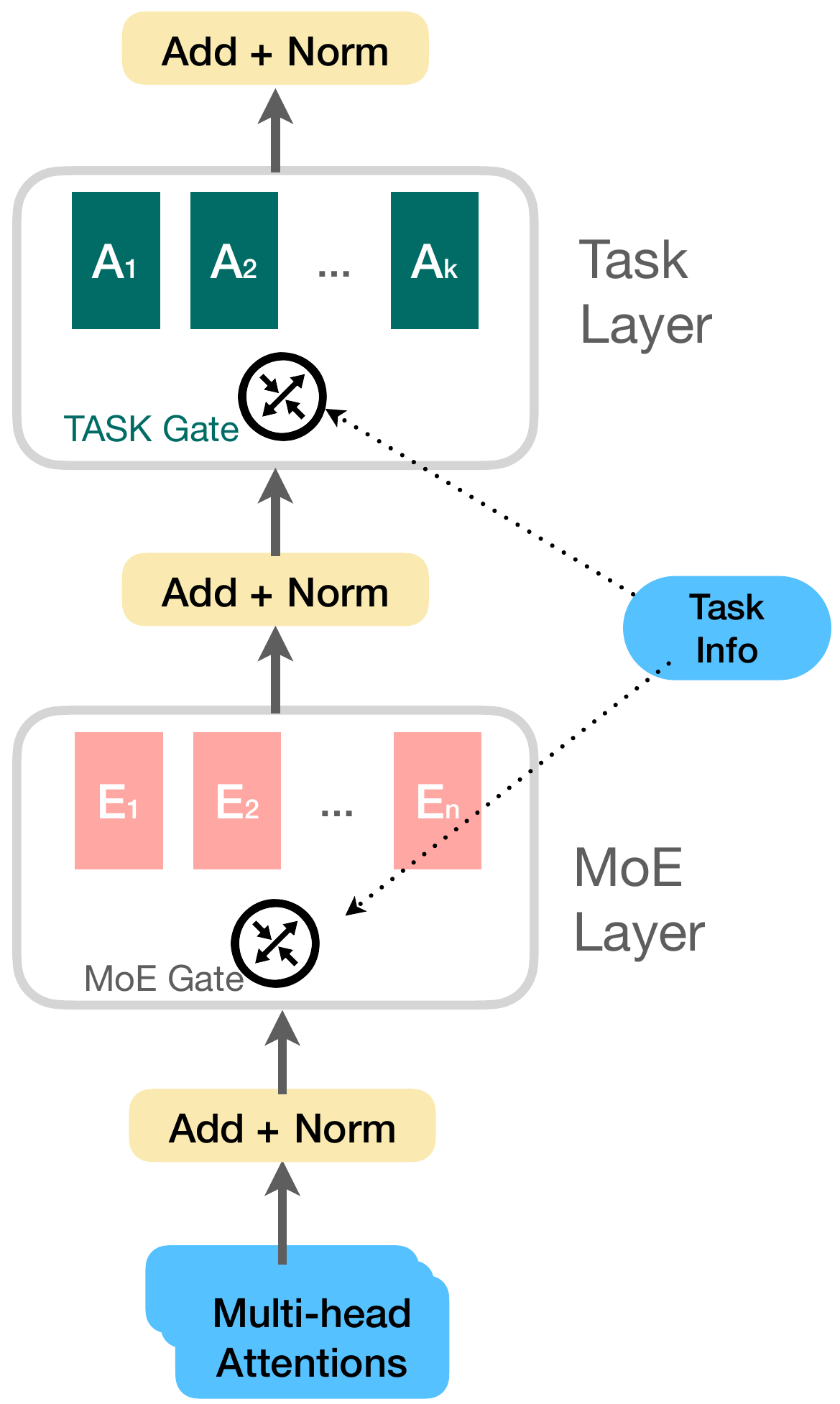}
  \caption{Extended from the typical MoE approaches that do not discriminate tokens from different tasks, we create shared task-related adapters that are trained to route tokens from similar tasks to the same shared adapters, and vice versa. 
   }
\end{figure} \label{fig:overview}

As a result, in this work, we design a novel MoE approach where task information is used during training and inference for assigning experts based on individual task information. 
The intuition is to make the training more task-aware so those similar tasks would be routed to the same group of experts and vice versa. From the architectural perspective, we incorporate high-level application-specific information with the system-level information to make the model become task-aware and hence have a better strategy in allocating experts based on the characteristics of distinct tasks, as also illustrated in Figure~\ref{fig:overview}.

Our proposed architecture allows for grouping experts based on the similarity of tasks, i.e. similar tasks should use a similar group of experts and otherwise for different tasks, by using shared-task adapters. Our design of putting those adapters on top of MoE layers allows for flexibility in future extensions: if we want the model to acquire new tasks while still having similar resources, we only finetune new adapters, and if we want to scale the hardware resources, e.g. for more speed,  we simply deal with MoE layers with such new resources.

Our experiments and analysis show the advantages of using task information in MoE architectures in multiple settings including multitask multilingual machine translations, as well as its generalization in few-shot learning. 
In summary, our contributions are as follows. 
\begin{itemize}
    \item First, we design novel MoE architectures that dynamically allocate experts based on task information in the context of multilingual multitask machine translation, with many variations.
    
    \item Second, we thoroughly study the pros and cons of our approaches in training from scratch, finetuning as well as transfer learning. 
    
    \item Third, we implement our models on top of well-proven infrastructres for practicality and scalability including \texttt{deepspeed}~\cite{Rasley2020DeepSpeedSO}, \texttt{fairseq}~\cite{Ott2019fairseqAF} and \texttt{transformer}~\cite{vaswani2017attention}. 
\end{itemize}

\section{Related Work} \label{sec:related}

\textbf{MoE Basic} \ 
Transformer-based Mixture-of-Experts (MoE) architecture essentially sparsifies transformer architecture by replacing the heavy feed-forward network (FFN) with a sparse MoE layer with top-1 or top-2 gates~\cite{shazeer2017outrageously}. However, increasing the number of experts does not simply increase the performance~\cite{fedus2021switch,clark2022unified}, many approaches have been proposed together to tackle the large-scale MoE deployment, such as in ~\cite{kim2021scalable}. In large-scale deployment, however, additional techniques should also be employed to battle with memory issues such as ``sharding'' experts~\cite{lepikhin2020gshard} or stabilizing the training~\cite{zoph2022designing}, since the models are often deployed on separate nodes that mainly used GPUs with limited memory. 
The architecture in this work inherits all of those techniques, and in addition incorporates task information into MoE routing, which in turn directs data into separate task adapters. This kind of routing is, however, hardware-agnostic, unlike some other work such as in~\cite{zheng2022alpa,chenta,xiongscomoe}. 


\textbf{MoE Routing Techniques} \ 
Gating is critical to MoE layer, which works as a weighted sum of the experts and serves for the ultimate purpose of load balancing of all available experts during both training and inference. Unlike the originally proposed top-k experts~\cite{shazeer2017outrageously,du2021glam}, it was studied in SwitchTransformer that a single expert can preserve the quality if chosen properly, while significantly reducing the communication and computation cost~\cite{fedus2021switch}. In more detail, SwitchTransformer first divides evenly amongst all experts with an optional buffer for imbalanced cases and then applies an auxiliary loss to enforce load balancing. Another alternative approach, which is more computationally efficient is to  get rid of such extra-heavy complicated loss and instead use a hash function to route every token to its matched expert, which tends to balance the output~\cite{roller2021hash}. Another interesting approach is to permit each token to appear in the top-k list of multiple experts~\cite{zhou2022mixture}, which has been proven to help, although not applicable for auto-regressive applications. 
Yet because of the inherent problem of load imbalance, another approach is to replace the gating mechanism with a stochastic selection method, which randomly activates an input during processing~\cite{zuo2021taming}. The intuition is somewhat similar to the hash approach since it relies on the ``fair'' randomness to solve the balance problem while keeping the blueprint more lightweight than enforcing an auxiliary loss. Along similar lines, research directions have explored the random dropping of outputs from MoE layers~\cite{liu2022gating,elbayad2022fixing}.
Unlike all of those routing techniques which are application agnostic, our proposed model connects the application level (i.e. task information) with the lower-level MoE layers for better dealing with interference of different tasks in the context of multilingual multitask applications.




\textbf{Task-level Routing} \
Recently task information has been used for improving MoE, e.g. in~\cite{zhilitask}. Our model is, however, much simpler and can be trained end-to-end unlike their approach, which requires clustering to be made for off-the-shelf shared representation learning. 
Probably the most related work to ours is Mod-Squad~\cite{Chen2022ModSquadDM} which shares the motivation with us while having several differences. First, their approach has multiple aids to make the task-based MoE work with an additional loss for regularization, while we instead rely mainly on the simple motivation of incorporating task information into MoE. 
Second, we still stick to a single gate for routing, while they allocate multiple gates, each \textit{per} task. 
Third, they additionally have MoE attention blocks, which make their architecture more complicated. 
Finally, our focused application is text-based machine translation, unlike computer vision settings in both works mentioned.

\section{Models} \label{sec:model}

Transformer architecture~\cite{vaswani2017attention} has been proven to be the core backbone of the pervasive successes in natural language processing, computer vision, and other artificial intelligence fields.
The main bottleneck to this architecture is, however, its heavy blueprint leading to intensive resources in training and inference, and is difficult to scale up. MoE is one powerful method to alleviate those problems in transformers.

\subsubsection{Sparse Mixture-of-Experts (MoE)} 
MoE, which was first introduced before the deep learning era~\cite{Jacobs1991AdaptiveMO}, was recently borrowed to address those drawbacks in transformer architecture~\cite{lepikhin2020gshard}. In a nutshell, MoE creates an ensemble of experts in multi-layer transformer blocks in place of a single expert, typically in the form of a feed-forward neural network (FFN) that is dense with many parameters. 

In terms of formality, given an original FFN layer called $\Tilde{E}$, we clone it into another layer containing a set of $N$ experts with exactly the same architecture $\{ E_i\}_{i=1}^N$. Likewise, the number of parameters for this particular layer is increased by a factor of $N$. 

The typical granular level of applying those experts in the context of natural language processing is the token level. Given a token $x$, its learned representation before MoE layer is a vector $\mathbf{x}$, its post-MoE output $\mathbf{y}$ is the weighted average of those experts' output 

\begin{align}
    o_i &= E_i(\mathbf{x})  \\
    \mathbf{y} &= \sum_{i=1}^N W_i o_i, \label{moe_eq:2}  
\end{align}

\noindent where $W_i$ is the weight (importance) of the corresponding expert $E_i$. 

The key to MoE power and its well-proven successes in tandem with transformer architecture is its sparsity design: only one or few experts are activated (i.e. having non-zero weight) at any point in time in spite of many more parameters just introduced due to the ensemble. Typically the component responsible for this sparsity is a gate that was co-trained with experts to route tokens to their target expert(s), and eventually assigns only a single or few non-zero weights across all experts \textit{per} token to its output $G(\mathbf{x})$ typically using softmax and top-k method

\begin{align}
    g_{out} &= \text{softmax} \left( W_g \mathbf{x} \right)    \\
    G(\mathbf{x}) &= \text{Top\_K} \left( g_{out} \right)
\end{align}

\noindent With $G(\mathbf{x})$ being a set of $K$ chosen experts, equation~\ref{moe_eq:2} becomes

\begin{align}
    \mathbf{y} &=  \sum_{i \in G(\mathbf{x}) } W_i o_i 
\end{align}

The main architectural problem with this design is its scalability: the memory will be quickly used up as we increase experts, given the limitation of current compute resources allocated to a single compute node in any distributed environment. GShard~\cite{lepikhin2020gshard} was born to fix this issue by trading the memory for communication: allocating each expert to a single node and only aggregating them when needed, e.g. gradient averaging in training or weight averaging when saving a model. This elegant design has unlocked MoE's unlimited scalability and practicality in enterprise-level deployments, especially with the following-up work in optimizing for better architecture in computation and communication, as mentioned in Section~\ref{sec:related}.

\begin{figure*}[t]
    \centering
  \includegraphics[width=\textwidth]{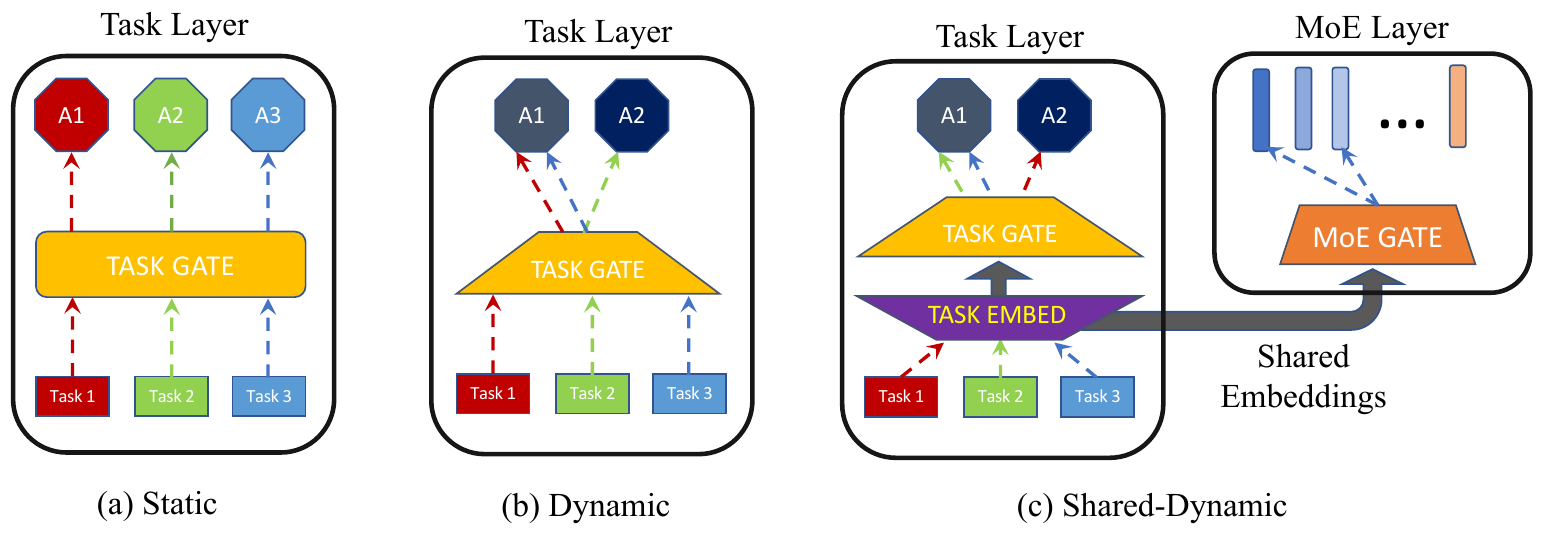}
  \caption{Our MoE models with variants. (a) \texttt{Static} means for each task, there is a separate adapter associated with it. (b) In the \texttt{Dynamic} mode, there is less number of adapters than the number of tasks, in order to learn the shared representation of similar tasks. (c) The last variant is \texttt{Shared-Dynamic} where the gates for task adapters and MoE share the same embedding for their routing decisions. 
  }
\label{fig:variants} 
\end{figure*}

\subsection{Task-based Adapters} \label{subsec:moe_task_adapters} 

Yet another problem on which we are focusing is not at the system level but more at the higher application level. As mentioned, in the multitask setting, the interference amongst tasks that are inherently different from each other could lead to the ineffectiveness of training. As a result, we employ \textit{task-based adapters} to separate those different tasks into different adapters. Likewise, data (or tokens) from similar tasks should be routed to a similar group of adapters. There are three modes for those adapters. 

First and the simplest is to allocate each adapter for each individual task. Although this setting is straightforward and requires no additional computation for data routing, it has the drawback of acquiring new unseen tasks. The reason is the model has to allocate a new adapter for each new task and fine-tune it with some amount of new data. Another potential problem is memory limitation if we want to extend to many new tasks in the future. This mode is called \texttt{static}, as shown in Figure~\ref{fig:variants}a. 

To enforce efficient learning of representation of similar task data, as well as alleviating memory problems, we have \texttt{dynamic} (Figure~\ref{fig:variants}b) where the number of adapters is less than the number of tasks. As a result, we intentionally guide the model to learn better cross-task representation in terms of similarity and dissimilarity. In other words, data from similar tasks should be routed to the same adapters and vice versa. In practice, we choose the number of adapters to be $\log_2(n)$ with $n$ being the number of tasks.

\subsection{Task-based Adapters with MoE} 

In this section, we formulate the task-based adapters mentioned in Section~\ref{subsec:moe_task_adapters} in combination with MoE, both of which are our core architecture components.  

Given $M$ tasks, we allocate them into $L$ shared-task adapters ($L < M$). For every single token $x$, we have the associated task information $t$ that makes up an input tuple $(\mathbf{x}, \mathbf{t})$ per token. As before, $\mathbf{x}$ is the representation vector from input, and $\mathbf{t}$ is the task representation vector learned by task embedding. 

Similar to MoE, we use a learnable task gate $G_t$ that is responsible for this routing with input being the concatenation of the input components

\begin{align}
    G_t(x, t) &= \text{Top\_K} (\mathbf{x} \oplus \mathbf{t})  \\
    \mathbf{y} &=  \sum_{i \in G_t(\mathbf{x}, \mathbf{t}) } W_i o_i 
\end{align}

And since the number of adapters $L < M$, the number of tasks, we call this setting \texttt{dynamic}, as demonstrated in Figure~\ref{fig:variants}b, as opposed to \texttt{static} (Figure~\ref{fig:variants}a), where each task will go to each individual adapter. 

Our main model uses the shared task embedding representation for the task gate as well as MoE gate, which we call \texttt{shared-dynamic}, as shown in Figure~\ref{fig:variants}c.





\section{Experiment Setup} \label{sec:experiment} 

\subsection{Data} \label{subsec:data} 
We tackle the problem of multitask multilingual machine translation using the data consisting of 10 different languages ranging from high-resource to low-resource ones including English (En), French (Fr), German (De), Czech (Cs), Finnish (Fi), Latvian (Lv), Estonian (Et), Romanian (Ro), Hindi (Hi), Turkish (Tr), and Gujarati (Gu). In more detail, the data for training, validation, and testing are listed in Table~\ref{tab:moe_data} where we can see besides the high-resource ones, we have low-resource languages such as Estonian, Hindi, or Gujarati.

\newcolumntype{K}[1]{>{\centering\arraybackslash}p{#1}}
\begin{table*}[t!]
\centering
\setlength\tabcolsep{1.0pt}
\begin{tabular}{c : c : *{9}{K{1.2cm}}}
\Xhline{3\arrayrulewidth} 
 & & \multicolumn{9}{c}{\textbf{Task Data}} \\
Split    &  Unit &  de-en & fr-en & cs-en & et-en & fi-en & gu-en & hi-en & lv-en & ro-en   \\
\Xhline{0.5\arrayrulewidth} 
Training & M & 4.6 & 10 & 10.3 & 0.7 & 4.8 & 0.9 & 0.3 & 1.4 & 0.5 \\
Validation & K & 3.0 & 3.0 & 3.0 & 2.0 & 1.4 & 2.0 & 0.5 & 2.0 & 2.0 \\
Testing & K & 3.0 & 3.0 & 3.0 & 2.0 & 1.4 & 2.0 & 0.5 & 2.0 & 2.0 \\
\Xhline{3\arrayrulewidth} 
\end{tabular}
\caption{Training, Validation, and Testing sizes for all XE tasks (the data for EX are exactly the same). Note that the unit for training is million (M) while that for both validation and testing are thousand (K), and the sizes are the same for validation and testing. 
}
\label{tab:moe_data}
\end{table*}

Those data are in the form of Bitext in which there is always English. As a result, we denote EX as the translation from English (E) to another language (X), and similarly for the other way around, XE. 
Those data are populated from the popular WMT corpus~\footnote{\url{https://www.statmt.org/wmt20/index.html}}. For the given 1 English and 9 other languages, there are consequently 9 EX and 9XE tasks. 
More information about data can be found in Table~\ref{tab:wmt} in Appendix~\ref{appx:wmt}.

\begin{table*}[h!]
\centering
\setlength\tabcolsep{1.0pt}
\begin{tabular}{c : *{9}{K{1.2cm}} : c }
\Xhline{3\arrayrulewidth} 
 & & \multicolumn{8}{c}{\textbf{{XE Tasks}}} \\
\textbf{Model}    &  de-en & fr-en & cs-en & et-en & fi-en & gu-en & hi-en & lv-en & ro-en  & \textbf{Avg} \\

\Xhline{0.5\arrayrulewidth}
1. Dense  & 29.9 & 31.2 & 28 & 22.4 & 21.4 & 22.3 & 21.4 & 24.5 & 36.1   & 26.4\\
2. MoE Token  & 27.9 & 29.5 & 26.3 & 19.9 & 19.6 & 18.9 & 17.7 & 22.3 & 33.8 & 24.0\\ 
3. MoE Sentence  & 27.9 & 29.9 & 26.2 & 21.4 & 19.9 & 17.9 & 15.9 & 23.2 & 34.4 & 24.1\\
4. MoE Task-Static & \textbf{32.1} & \textbf{33.3} & \textbf{30.7} & \textbf{24.3} & \textbf{23.4} & 20.6 & \textbf{22.5} & \textbf{27.2}  & \textbf{38.8} & \textbf{28.1} \\
5.MoE Task-Dynamic & \textbf{31.4} & \textbf{32.0} & \textbf{29.1} & \textbf{23.4} & \textbf{22.1} & 18.9 & 20.5 & \textbf{25.5} & \textbf{37.2} & \textbf{26.7} \\
\Xhline{0.5\arrayrulewidth}
\hline
& & \multicolumn{8}{c}{\textbf{EX Tasks}} \\
 & en-de & en-fr & en-cs & en-et & en-fi & en-gu & en-hi & en-lv & en-ro & \\
\Xhline{0.5\arrayrulewidth}
1. Dense  & 25.4 & 28.3 & 22.4 & 23.3 & 20.9 & 28.4 & 29.0 & 26.5 & 31.5 & 26.2\\
2. MoE Token & 22.9 & 25.1 & 19.5 & 20.1 & 17.9 & 26.2 & 26.3 & 24.0 & 29.0  & 23.4\\
3. MoE Sentence  &  23.2 & 25.7 & 20.4 & 22.4 & 18.7 & 26.4 & 27.1 & 24.2 & 29.7  & 24.2 \\
4. MoE Task-Static & \textbf{29.5} & \textbf{32.5} & \textbf{27.9} & \textbf{27.4} & \textbf{25.8} & \textbf{28.8} & \textbf{30.8} & \textbf{32.2} & \textbf{34.6}  & \textbf{29.9} \\
5.MoE Task-Dynamic & \textbf{27.3} & \textbf{29.6} & \textbf{25.0} & \textbf{24.7} & \textbf{22.7} & 27.7 & 2\textbf{9.3} & \textbf{28.4} & \textbf{32.7}  & \textbf{27.5} \\
\Xhline{3\arrayrulewidth}
\end{tabular}
\caption{Comparison of task-based MoE models (models 4 \& 5) to task-agnostic MoE models (models 2 \& 3) and the non-MOE (Dense) model (model 1) in BLEU scores.  
With the help of task information, task-based MoE models show their outperforming BLEU scores over all other types across most of the tasks including both high-resource and low-resource ones. 
}
\label{tab:moe_main}
\end{table*}

\subsection{Task and Model Training} \label{subsection:moe_model_and_train} 
In this section, we describe the task information, evaluation metrics, and how we deal with data and models for training. 

\textbf{Task} \, Our task is multitask multilingual machine translation (MMMT) which uses the EX and XE pairs. Our single model is trained with two main capacities. 
First, this single model can translate all the training pairs with high accuracy. 
Second, the model is able to quickly acquire new translation pairs with only zero or a few shots.  

\textbf{Evaluation} \, While there are many evaluation metrics, we mainly use BLEU score due to its popularity and credibility in evaluating machine translation tasks. This evaluation is implemented by SacreBLEU\footnote{\url{https://github.com/mjpost/sacrebleu}}. We note that, unlike all available public implementations that we found, our implementation evaluates all BLEU scores on the fly along with the training, so there is no disruption for offline evaluation. That also helps in early stopping based on the BLEU scores on the validation sets. 

\textbf{Pre-Processing and Post-Processing} \ 
In terms of preprocessing, we first encode the data using the Byte-Pair encoding (BPE) method and generate shared dictionaries where all the language pairs use the same vocabulary of size 64K, before feeding to the model. To get accurate scores, for post-processing, we again use BPE decoding for reconstructing the whole translated sentences before comparing them with the original sentences before BPE pre-processing. Likewise, we treat all the processing and model manipulation as a black box for calculating the scores. 

\textbf{Model Configuration and Implementation} \ 
We use transformer architecture~\cite{vaswani2017attention} with 12 layers for both encoder and decoder phases, each of which uses a word embedding layer of dimension 1024 and a non-linear layer of dimension 4096. There are 16 attention heads and a dropout rate of 30\%. For MoE, all jobs are trained on Azure cloud machines with 8 GPUs, each of which takes around 2 weeks for a model covering 18 aforementioned tasks to reach decent scores. We apply early stopping based on the validation BLEU scores, in which a non-increasing score after 2 epochs is the condition.  For task-based information, we have a task embedding dimension of 64 and a task adapter hidden dimension of 256 for every single task adapter. Our implementation inherits the lower-level infrastructure code from Microsoft Deepspeed and Fairseq.~\footnote{\url{https://github.com/facebookresearch/fairseq}}

As for the implementation, an important practical issue with MoE is load balancing among experts for the best utilization of the infrastructure systems. For enforcing the training to have a balanced load, as a result, we employ the auxiliary loss from~\citet{lepikhin2020gshard}. 

\subsubsection{Baselines}
In order to show the performance of the task-based MoE models, the following baselines are selected: 

\textbf{Dense} \ This is the traditional transformer model without any MoE layer, i.e., no change to the fully connected (FFN) layer in each layer of encoders or decoders. 

\textbf{MoE - Token} \ This is the MoE model that is usually considered the default option where each FFN layer is replaced by an MoE layer. In our experiments, each MoE layer comprises 8 experts (each has the same size as the original FFN being replaced) and a gate for routing purposes. 

\textbf{MoE - Sentence} \ This is yet another MoE architecture with exactly the same architecture configuration as the MoE - Token baseline. The difference is in the routing layer, which functions at a different granularity: sentences instead of tokens. In more detail, while the gate decides which expert for each token separately in MoE - Token model, it instead routes all tokens belonging to a single sentence to the same chosen expert.

\section{Results and Discussions} \label{sec:moe_result}

\begin{table*}[]
\centering 
\setlength\tabcolsep{1.1pt}
\begin{tabular}{c:cc:cc:cccc:c}
\Xhline{3\arrayrulewidth} 
\multirow{2}{*}{\textbf{Model}} & \multicolumn{2}{c}{\textbf{Design}}     & \multicolumn{2}{c}{\textbf{Routing}}   & \multicolumn{4}{c}{\textbf{Tasks}}                                & \multirow{2}{*}{\textbf{Average}} \\
\cmidrule(lr{.5em}){2-3} \cmidrule(lr{.5em}){4-5} \cmidrule(lr{.5em}){6-9}
                                & MoE |                & Task               & MoE                   & Task           & {de-en} & {fr-en} & {et-en} & {fi-en} &                                   \\
               
\Xhline{0.5\arrayrulewidth} 
MoE                             & Y                  & N                  & Token                 & -              & 32.4           & 33.7           & 24.2           & 23.6           & 28.5                              \\
\cmidrule(lr{.5em}){2-3} \cmidrule(lr{.5em}){4-5}
Dense + Task Static             & \multirow{2}{*}{N} & \multirow{2}{*}{Y} & \multirow{5}{*}{Task} & Static         & 32.2           & 33.7           & 21.0             & 22.8           & 27.4                              \\
Dense + Task Dynamic            &                    &                    &                       & Dynamic        & 31.9           & 33.0             & 22.0             & 22.5           & 27.4                              \\
\cmidrule(lr{.5em}){2-3} \cmidrule(lr{.5em}){5-5}
MoE + Task Static               & \multirow{3}{*}{Y} & \multirow{3}{*}{Y} &                       & Static         & 30.7           & 32.0             & 19.9           & 20.8           & 25.9                              \\
MoE + Task Dynamic              &                    &                    &                       & Dynamic        & \textbf{32.6}  & \textbf{33.9}  & 24.0             & \textbf{23.9}  & \textbf{28.6}                     \\
MoE + Task Shared-Dynamic       &                    &                    &                       & Shared-Dynamic & 32.2           & 33.3           & \textbf{24.3}  & \textbf{24.5}  & \textbf{28.6}                                 \\   
\Xhline{3\arrayrulewidth} 
\end{tabular}
\caption{Performance of different models with changes on whether MoE layers exist, whether Task Adapters exist, and how routing for those components is undertaken. The scores better than the baseline are highlighted. Task-based MoE shows advantages, especially with shared-dynamic adapters between MoE and Task Adapters on the low-resource language pair.  
}
\label{tab:moe_ablation}
\end{table*}

\subsection{Multitask Multilingual Machine Translation} 
We first present the main results for models capable of translating 18 tasks (see Section~\ref{subsection:moe_model_and_train}) concurrently. 
As shown in Table~\ref{tab:moe_main}, our models that incorporate MoE layers and are enhanced with task information show great advantages over all the baseline models on most tasks, in both directions EX and XE, in accordance with our hypothesis that using task adapters in conjunction with MoE is helpful in multilingual multitask translation. 

An outstanding drawback with which the task-based MoE models are facing, however, is for the low-resource translation pairs, e.g. Gu-En, Hi-En, or En-Gu. As we can see from the results in Table~\ref{tab:moe_main}, training those pairs with Dense models seems to benefit more than with MoE models. We hypothesize the problem is due to the undersampling of the training data for those languages, which have much less data than their high-resource counterparts. 
In more detail, our training routine concatenates all the tasks' data in a single big dataset before drawing batches. However, without adjusting the sampling process, high-resource language pairs are being trained significantly more given their much larger data in place. 
In particular, for the case of Gujarati where the Task-Dynamic MoE model underperforms in comparison to the baselines, our hypothesis is that linguistically, this language is the most different from all other languages, which makes the models very hard to learn effective shared representation with any other pairs. 

In the future, we plan to explore ideas such as custom sampling or contrastive representation learning to tackle with such issues with the low-resource language pairs, in order to make MoE work as well for those languages as in high-resource pairs.

\begin{figure*}[ht!]
\begin{subfigure}{.3\textwidth}
  \centering
  \includegraphics[width=0.95\linewidth]{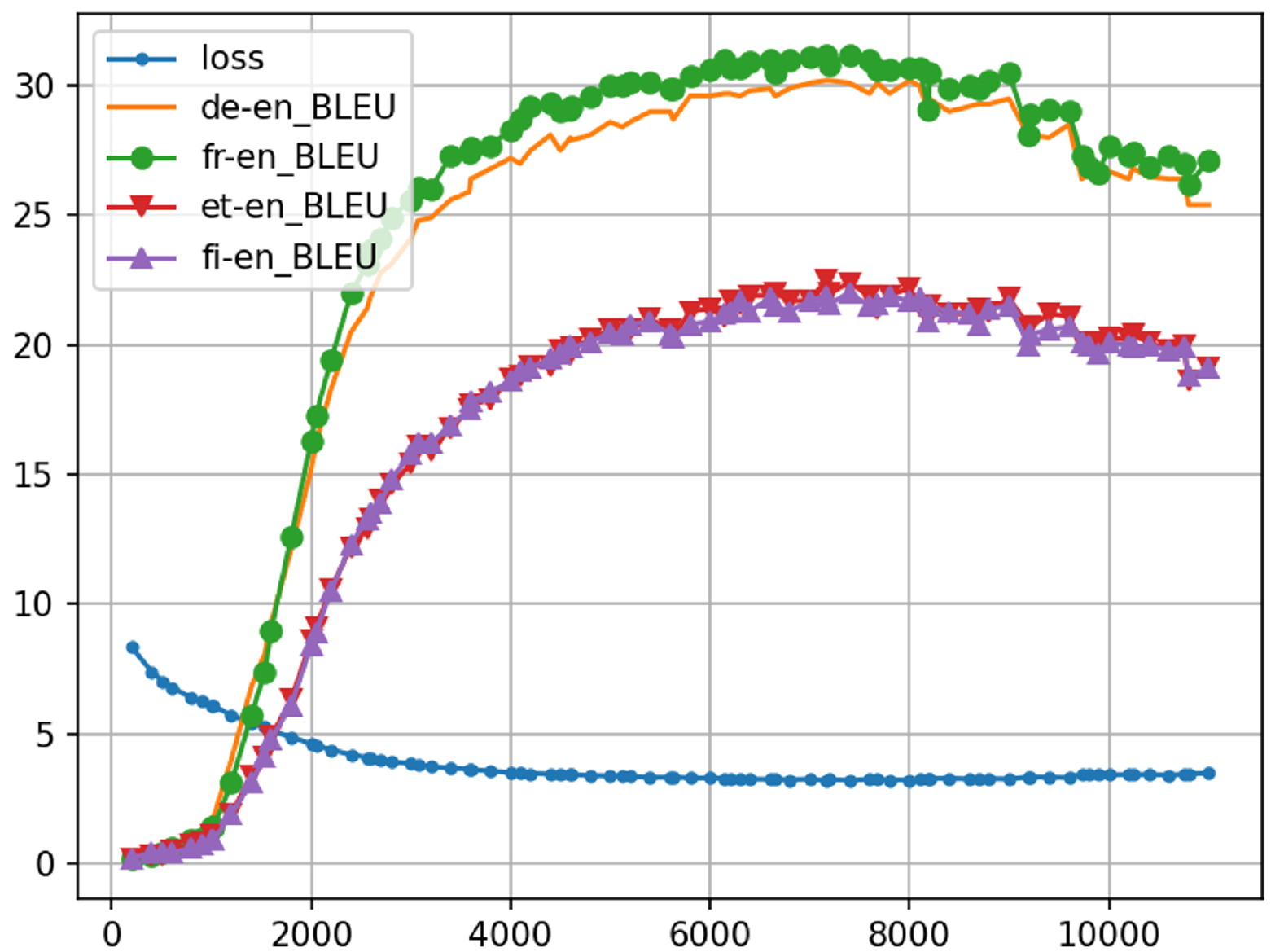}  
  \caption{model 1}
  \label{fig:sub-first}
\end{subfigure}
\begin{subfigure}{.3\textwidth}
  \centering
  \includegraphics[width=0.95\linewidth]{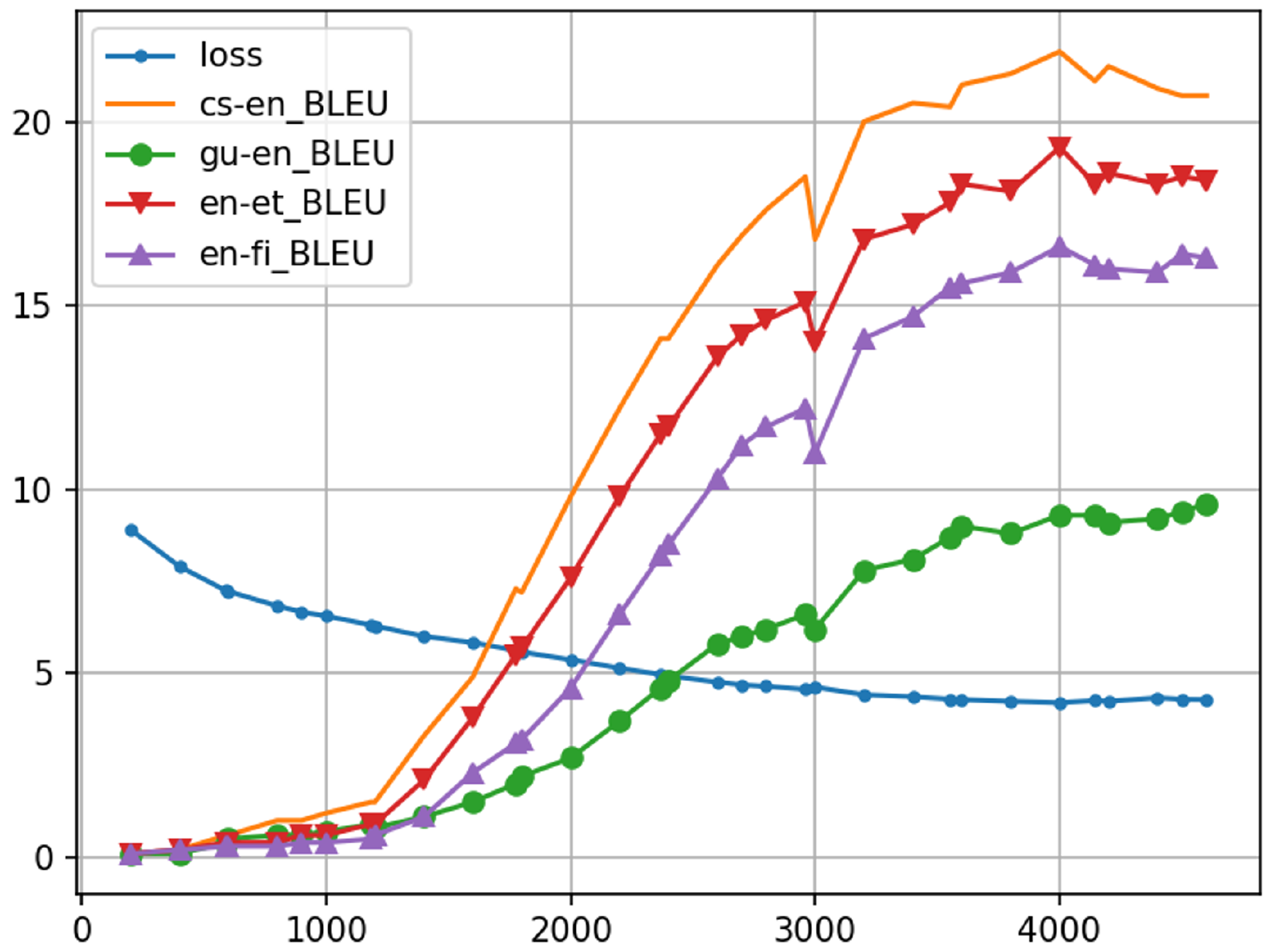}  
  \caption{model 2}
  \label{fig:sub-second}
\end{subfigure}
\begin{subfigure}{.41\textwidth}
  \centering
  \includegraphics[width=0.95\linewidth]{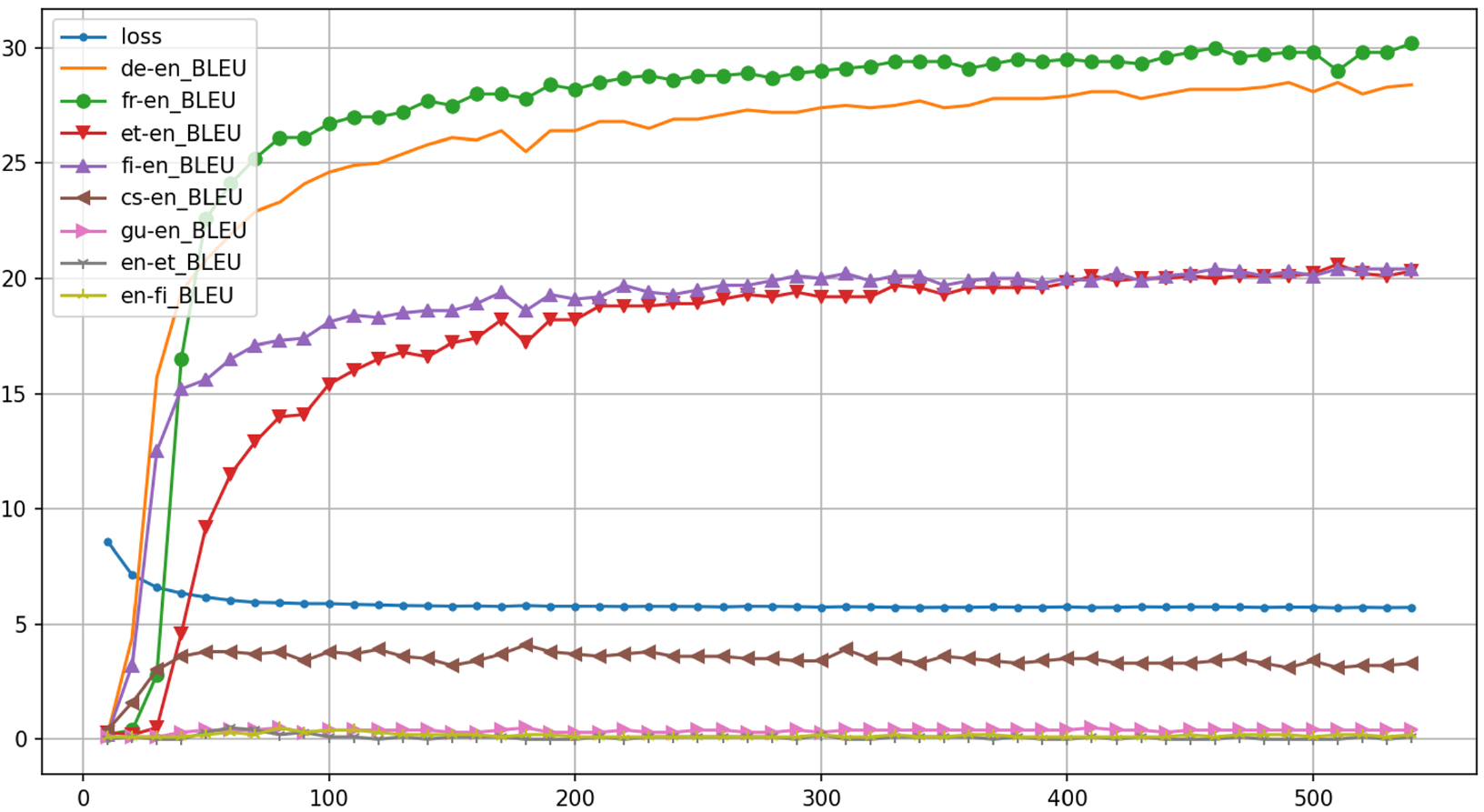}  
  \caption{merged model}
  \label{fig:sub-merged}
\end{subfigure}
\caption{Ablation study about merging 2 checkpointed models of different capabilities. Model 1 is trained with 4 tasks: de-en, fr-en, et-en and fi-en. Model 2 is trained with the other 4 tasks: cs-en, gu-en, en-et, and en-fi. Although those 2 models are under-trained with only a few thousand steps, in the merged model that has the capabilities of those two combined, many pairs have quickly picked up to a similar levels as in the previous single models.} 
\label{fig:merged}
\vspace*{-3mm}
\end{figure*}

\subsection{Ablation Study}

\subsubsection{Implications of Different Task Layers and MoE Layers} \label{subsubsec:diff_ablation}

In this study, we limit the number of tasks to four (De-En, Fr-En, Et-En, and Fi-En), which can be divided into 2 groups of similar tasks: (De-En, Fr-En) is the first group and (Et-En, Fi-En) is the second one, to study the performance implications of different model variants when there is a task layer and/or MoE layer. 

As illustrated in Table~\ref{tab:moe_ablation}, we again see that combining MoE and Task Adapters yields the best models, the same trend as shown in Table~\ref{tab:moe_main}, particularly when the dynamic adapters are used to enforce similar tasks to share the same representations. 

However, when task adapters are not used in conjunction with MoE, the performance is worse than MoE alone. This also means MoE should be the foundational infrastructure, and on top of that, task adapters should be used. It concurs with the motivation that the interference of different tasks or languages makes the training of experts difficult. In other words, there is not so much help when there is only one expert for all the tasks (i.e. Dense models).

\subsubsection{Flexibility of Task-based MoE in Merging Two Trained Models}

One of the important capabilities in multitask learning and in general learning problems is how to quickly acquire new capabilities given current models with minimal resources and effort. 
Aligned with this goal, this ablation explores how quickly our task-based MoE models can be merged with each other from 2 different models to newly establish only 1 model that has the combination of their capabilities. 

In merging those two models, we restore two respective checkpoints and merge layer-by-layer as follows. 
First, task-based adapters are kept and combined with each other: each model has 2 adapters (for 4 tasks in the model) and the combined model has 4 adapters (for 8 tasks in combination). Second, the task routers will also be merged and changed so that the routing of each data will now have 4 selections instead of 2 outputs as in the previous models. Finally, the rest of the transformer and MoE layers will have their weights averaged. 

The tasks in the original two models are hand-picked as in Section~\ref{subsubsec:diff_ablation} to have 2 different groups, each of which has 2 similar tasks. Model 1 has de-en, fr-en, et-en, and fi-en, while Model 2 has cs-en, gu-en, en-et and en-fi.

As shown in Figure~\ref{fig:merged}, while two original models have been trained with just a few thousand steps (a couple of hours), the combined model shows that it can quickly pick up their original capabilities with just a few hundred steps after merging. Although there are a few uncommon pairs that seem to fail, such as gu-en or en-et, the chart shows the optimistic result of combining trained models with our flexible task-based MoE architectures. 

\section{Conclusion} 

In the era of large language models, more efficient and effective modeling techniques are essential to, where MoE in combination with transformer-based models has proven its great advantages. It is, however, complicated to enable that implementation in practice due to the difficulties of training a single model serving diverse tasks. The proposed task-based MoE, which employs both task adapters in tandem with MoE has shown its promising advantages in the application of multitask multilingual machine translations. This novel design enforces shared representation of similar tasks and separates different task data to counter the interference effects. In addition, it also offers the flexibility of changing adapters based on new tasks or changing the MoE infrastructure without affecting the application level. 
Besides outperforming the traditional approaches using Dense models, however, our MoE models still need to improve on low-resource language pairs. 
To tackle that issue, in the future, exploring custom sampling for those pairs, and enforcing the shared representation learning explicitly using such additional techniques as contrastive learning or mutual information are worth exploring. 

\section{Acknowledgements}

The authors would like to thank the great feedback and help from Yiren Wang, Muhammad ElNokrashy, Alex Muzio, Akiko Eriguchi and other members of Microsoft's Machine Translation Group. 

\bibliography{mybib}


\appendix 

\section{WMT Data Information} \label{appx:wmt} 
\begin{table}[ht!]
\centering
\setlength\tabcolsep{4.2pt}
\begin{tabular}{ccc}
\Xhline{3\arrayrulewidth} 
Code & Language & Test Split    \\
\Xhline{0.5\arrayrulewidth} 
de   & German   & wmt2013       \\
fr   & French   & wmt2013       \\
cs & Czech & wmt2013 \\
et & Estonian & wmt2018dev \\ 
fi & Finish & wmt2015 \\ 
gu & Gujarati & wmt2019dev \\ 
hi & Hindi & wmt2014dev \\ 
lv & Latvian & wmt2017dev \\ 
ro & Romanian & wmt2016dev \\
\Xhline{3\arrayrulewidth} 
\end{tabular}
\caption{More details about our datasets for comparison and reproducibility. 
}
\label{tab:wmt}
\end{table}

\end{document}

%% file: math_commands.tex
\usepackage{amsmath,amsfonts,bm}
\usepackage{amsthm}
\usepackage{mathtools}

        {\hspace*{\fill}$\Box$\par}

\DeclarePairedDelimiterX{\infdivx}[2]{(}{)}{%
  #1\;\delimsize\|\;#2%
}

\def\eqref#1{equation~\ref{#1}}

\def\1{\bm{1}}

\DeclareMathAlphabet{\mathsfit}{\encodingdefault}{\sfdefault}{m}{sl}
\SetMathAlphabet{\mathsfit}{bold}{\encodingdefault}{\sfdefault}{bx}{n}

